\documentclass[final]{article}

\usepackage{arxiv}
\usepackage[utf8]{inputenc} 
\usepackage[T1]{fontenc}    
\usepackage{hyperref}       
\usepackage{url}            
\usepackage{booktabs}       
\usepackage{amsfonts}       
\usepackage{nicefrac}       
\usepackage{microtype}      

\usepackage{lipsum}
\usepackage{graphicx}
\usepackage{authblk}
\usepackage{amssymb}
\usepackage{latexsym}
\usepackage{amsmath}
\usepackage{amsfonts}
\usepackage{mathrsfs}

\usepackage{algorithm}
\usepackage{algorithmic}
\usepackage{graphicx}
\usepackage{afterpage}
\usepackage{wrapfig}
\usepackage{bbm}
\usepackage{array}
\usepackage{booktabs}
\usepackage{caption}
\usepackage[table]{xcolor}
\usepackage{url}
\usepackage{hyperref}
\usepackage{lipsum}
\usepackage{CJKutf8}
\usepackage{multirow}
\usepackage{multicol}

\usepackage{lineno} 

\usepackage{float}
\usepackage{pdfpages}
\usepackage{changes}  
\usepackage{xcolor}
\usepackage{pifont}

\newcommand{\ie}{{\textit{i.e.}}}
\newcommand{\etal}{{\textit{et al.}}}
\newcommand{\eg}{{\textit{e.g.}}}
\newcommand{\etc}{{\textit{etc.}}}

\definechangesauthor[name={Author1}, color=red]{A1} 
\definechangesauthor[name={Author2}, color=green]{A2} 

\definecolor{newcolor}{rgb}{.8,.349,.1}

\title{Towards Unifying Understanding and Generation in the Era of Vision Foundation Models: A Survey from the Autoregression Perspective}

\author{
  \textbf{Shenghao Xie}$^{1\ast}$, 
  \textbf{Wenqiang Zu}$^{2,4\ast}$, 
  \textbf{Mingyang Zhao}$^{2\ast}$,
  \textbf{Duo Su}$^{3}$,
  \textbf{Shilong Liu}$^{3}$,
  \textbf{Ruohua Shi}$^{1}$,\\ 
  \textbf{Guoqi Li}$^{2}$, 
  \textbf{Shanghang Zhang}$^{1,4\dag}$,
  \textbf{Lei Ma}$^{1,4\dag}$
}

\affil{$^1$ Peking University}
\affil{$^2$ Chinese Academy of Sciences}
\affil{$^3$ Tsinghua University}
\affil{$^4$ Beijing Academy of Artificial Intelligence}

\begin{document}
\maketitle

\renewcommand{\thefootnote}{\fnsymbol{footnote}}
\footnotetext[1]{Equal contribution. Emails: \textit{shenghaoxie@stu.pku.edu.cn}.}
\footnotetext[2]{Corresponding authors. Emails: \textit{lei.ma@pku.edu.cn}, \textit{shanghang@pku.edu.cn}.}

\begin{abstract}\label{sec: 0_abstract}
  Autoregression in large language models (LLMs) has shown impressive scalability by unifying all language tasks into the \textit{next token prediction} paradigm. Recently, there is a growing interest in extending this success to vision foundation models. In this survey, we review the recent advances and discuss future directions for autoregressive vision foundation models. First, we present the trend for next generation of vision foundation models, \ie, unifying both understanding and generation in vision tasks. We then analyze the limitations of existing vision foundation models, and present a formal definition of autoregression with its advantages. Later, we categorize autoregressive vision foundation models from their vision tokenizers and autoregression backbones. Finally, we discuss several promising research challenges and directions. To the best of our knowledge, this is the first survey to comprehensively summarize autoregressive vision foundation models under the trend of unifying understanding and generation. A collection of related resources is available at \url{https://github.com/EmmaSRH/ARVFM}.
\end{abstract}
\section{Introduction} \label{sec: 1_introduction}
Foundation models~\cite{bommasani2021foundationmodel} have achieved remarkable success in the field of natural language processing (NLP)~\cite{chowdhary2020nlp}, \eg, GPT~\cite{radford2018gpt}, Llama~\cite{touvron2023llama}, and Qwen~\cite{bai2023qwen}. As the size of datasets and model parameters increases, they demonstrate robust and generalizable performance in downstream tasks, \ie, scaling law~\cite{kaplan2020scalinglaw}. This is largely attributed to advancements in computing hardwares, \eg, GPU, the emergence of parallel architectures, \eg, Transformer~\cite{vaswani2017transformer}, and the increasing of training data. In particular, large-scale available data is greatly benefiting from the autoregression paradigm adopted by LLMs, which unifies most language downstream tasks, \eg, text classification~\cite{kowsari2019textclassification}, sentiment analysis~\cite{medhat2014sentimentanalysis}, and question answering~\cite{allam2012questionanswering}, into only one task: generating the next token based on previous ones, \ie, next token prediction. This further bridges the gap between different tasks, allowing data from various domains to be integrated together, and significantly enhancing the generalization capability of models.

Generally, Yuan \etal~\cite{yuan2021visionfoundationmodel} suggest that the vision foundation model is a pretrained model with its adapters for addressing all vision tasks, \ie, both understanding and generation,  within the \textit{Space-Time-Modality} space, and ensuring transferability. This space encompasses three dimensions: 1) \textit{Space}: from coarse-grained to fine-grained entities, \eg, scene and object. 2) \textit{Time}: from spatial to temporal dimension, \eg, 2D/3D image and video. 3) \textit{Modality}: from RGB to other formats, \eg, caption and depth map. However, compared to language tasks, where both input and output spaces can naturally be treated as 1D word sequences, the complexity of \textit{Space-Time-Modality} space makes it challenging to unify vision tasks. Therefore, most previous vision foundation models are still designed for specific tasks, \eg, SAM~\cite{kirillov2023sam}, Depth Anything~\cite{yang2024depthanything}, and Recognize Anything~\cite{zhang2024recognizeanything}. Specifically, we can observe that previous vision foundation models usually focus separately on understanding or generation, \eg, CLIP~\cite{radford2021clip}, BLIP~\cite{li2022blip}, LLaVA~\cite{liu2024llava}, and Diffusion~\cite{ho2020ddpm}. In this way, Dong \etal~\cite{dong2024dreamllm} point out that the synergy between understanding and generation is heavily overlooked, and the conditional, marginal, and joint distributions of the vision modality is modeled incompletely. Therefore, the first significant challenge is: \textit{How to find an effective and flexible approach to unify all vision tasks, especially in achieving a unified framework for understanding and generation?} Inspired by the in-context learning~\cite{dong2022icl} in NLP, the most intuitive way is to specify vision tasks through natural language~\cite{liu2023groundingdino, shen2024hugginggpt}. Later, the concept of \textit{images speak in images} is considered more suitable for designing instructions that address a boarder range of vision tasks~\cite{wang2023painter}, and the visual sentence~\cite{bai2024lvm} is proposed to enable the model to infer the task to be performed on the query image based on input refrence images and their ground truth \eg, images with annotation, 3D rotation viewpoints, and video sequences. Although we can now preliminarily specify any vision task in terms of input format, mapping the input \textit{Space-Time-Modality} space into a shared intermediate space for joint representation learning still remains challenging. Following NLP, the token-based representation is introduced, \ie, visual token~\cite{peng2024visualtoken, dosovitskiy2020vit}. By dividing different vision inputs in the \textit{Space-Time-Modality} space into a series of tokens and mapping them into vision embeddings based on specific rules, we can progressively solve all vision tasks similarly to LLMs. The vision tokenizer is the key component to accomplish above functions.

Moreover, existing research~\cite{xie2023maescalinglaw, xiao2021clscalinglaw1, wang2022clscalinglaw2} finds that learning paradigms commonly used in previous vision foundation models, \eg, masked image modeling (MIM)~\cite{he2022mae} and contrastive learning (CL)~\cite{he2020moco}, may face bottlenecks in further exploiting the scalability of Transformer. This may be due to a difference between their pretraining objectives, \eg, reconstructing the masked region in MIM, and the following downstream task objectives, \eg, image classification~\cite{krizhevsky2017alexnet}, potentially causing conflicts between the signals obtained from self-supervised learning and those required for downstream tasks. Therefore, the second significant challenge is: \textit{How to find an effective paradigm to process vision tokens in an end-to-end manner, with a straightforward and consistent training objective?} Inspired by LLMs, autoregression can learn effective representations from large-scale data manifolds by simply predicting the next token, and fully leverages the scalability of Transformer. The autoregression Transformer architecture has been widely adopted as the backbone in recent vision foundation models, \ie, autoregressive vision foundation models.         

This paper aims to provide a comprehensive survey of recent advancements and future directions of autoregressive vision foundation models under the trend of unifying understanding and generation. In Sec.~\ref{sec: 2_preliminary}, we first illustate our target scope of vision foundation models with downstream tasks. We then define the concept of autoregression. Then, in Sec.~\ref{sec: 3_method}, we present a detailed taxonomy and an in-depth analysis of the vision tokenizers and autoregression Transformer backbones. In Sec.~\ref{sec: 4_discussion}, we continue to introduce the potential research challenges and directions of autoregressive vision foundation models. In Sec.~\ref{sec: 5_conclusion}, we finally conclude this survey.   
\section{Preliminary}\label{sec: 2_preliminary}
\paragraph{Vision Understanding and Generation.} Typically, the downstream vision tasks are divided into two categories: understanding and generation. Foundation models for learning vision understanding through language posteriors have both made significant progress, but they lack vision posteriors~\cite{dong2024dreamllm}. This is still far from building a general vision foundation model. In the \textit{Space-Time-Modality} space, vision understanding includes not only tasks like image captioning~\cite{wang2020imagecaption}, retrival~\cite{chen2021imageretrival}, and classification \etc, through text, but also segmentation~\cite{minaee2021segmentation}, detection~\cite{liu2023groundingdino} and tracking~\cite{smeulders2013visualtracking} \etc~Similarly, vision generation includes not only translation, \eg, text to image~\cite{ding2021cogview}, but also tasks like denoising~\cite{fan2019denoise}, super-resolution~\cite{dong2015superresolution}, and enhancement~\cite{qi2021enhancement} \etc~Recent studies~\cite{dong2024dreamllm} suggest that generation and understanding have a synergistic effect and mutually enhance each other. Furthermore, with advancements in token-based vision representation and in-context learning, the unification of vision tasks becomes possible.
\textit{Both vision understanding and generation can be unified as a generation task, \ie, generating tokens in the \textit{Space-Time-Modality} space}. Autoregression has become the mainstream paradigm for accomplishing this task. Therefore, in this survey, we focus primarily on autoregressive vision foundation models, which can further be described as \textit{autoregressive generative vision foundation models}. In such scope, these models have significant potential to comprehensively model the \textit{Space-Time-Modality} space for both input and output, achieving the unification of all vision tasks, \ie, both understanding and generation, in the future.

\paragraph{Autoregression.} Following AIM~\cite{el-nouby2024aim}, we give the basic knowledge of autoregression. In autoregressive vision foundation models, images are tokenized as a sequence of tokens $x=(x_1, x_2, \ldots, x_K)$, where $K$ is the product of the height and width of an image feature map and also the length of vision tokens.
Then, the next-token prediction is defined as
\begin{equation}
p\left(x_1, x_2, \ldots, x_K\right)=\prod_{k=1}^K p\left(x_k \mid x_{<k}\right),
\end{equation} 
where $p$ represents the probability. 
Each token is predicted only depending on their prefix tokens $x_{<k}=(x_1, x_2, \ldots, x_{k-1})$.
Finally, the predicted token sequence is decoded into image space.

For a given set of images $x \in\mathcal{X}$, the training objective of autoregresion is to minimize the following negative log-likelihood:
\begin{equation}
\sum_{k=1}^K-\log p\left(x_k \mid x_{<k}\right) .
\end{equation}
Under the Gaussian distribution assumption, we can instantiate the training objective with the regression loss, typically using a distance metric, \eg, the Euclidean distance:
\begin{equation}
\mathcal{L}=\min _\theta \frac{1}{K} \sum_{k=1}^K\left\|\hat{x}_k(\theta)-x_k\right\|_2^2.
\end{equation}
In our survey, we take the analysis of vision tokens as an example, there are also scenarios for jointly modeling tokens from other modalities, \eg, text and audio.

\paragraph{Comparison with Related Survey.} Although existing surveys~\cite{liu2024survey1, li2024survey2} have also acknowledged the latest trend of unifying both understanding and generation, they have not continued to provide a detailed and comprehensive discussion on the underlying mainstream paradigms and related works under this trend, \ie, autoregressive vision foundation models. We also present a deeper insight about unifying vision understanding and generation. Azim \etal~\cite{azimsurvey2} present the application of autoregressive models in image and video generation, but mainly focus on small models instead of foundation models, and do not show the significance of autoregression in the context of unifying vision understanding and generation. To the best of our knowledge, this is the first survey to comprehensively summarize autoregressive vision foundation models under the trend of unifying understanding and generation.

\begin{table*}[!t]
\centering
\setlength{\tabcolsep}{6pt}
\small
\begin{tabular}{
    >{\centering\arraybackslash}p{3cm}
    >{\centering\arraybackslash}p{2cm}
    >{\centering\arraybackslash}p{2cm}
    >{\centering\arraybackslash}p{2cm}
    >{\centering\arraybackslash}p{2cm}
    >{\centering\arraybackslash}p{1cm}
}
\toprule
 \textbf{Model} & \textbf{Tokenizer} & \textbf{Backbone} & \textbf{Publication} & \textbf{Code} \\
\midrule
Fluid~\cite{fan2024fluid}& LQ & Bidirectional & arXiv 2024 & \href{https://github.com/rongyaofang/PUMA}{\checkmark} \\

PUMA~\cite{fang2024puma}& LQ & Causal & arXiv 2024 & \href{https://github.com/rongyaofang/PUMA}{\checkmark} \\

Janus~\cite{wu2024janus}& DP+VQ & Causal & arXiv 2024 & \href{https://github.com/deepseek-ai/Janus}{\checkmark} \\

SAR~\cite{liu2024sar}& VQ & Causal+Bidirectional & arXiv 2024 & \ding{55} \\

HART~\cite{tang2024hart}& VQ+LQ & Causal & arXiv 2024 & \href{https://github.com/mit-han-lab/hart}{\checkmark} \\

MMAR~\cite{yang2024mmar}& LQ & Bidirectional & arXiv 2024 & \href{https://github.com/ydcUstc/MMAR}{\checkmark} \\

Baichuan-Omni~\cite{li2024baichuanomni}& DP & Causal & arXiv 2024 & \href{https://github.com/westlake-baichuan-mllm/bc-omni}{\checkmark} \\

ARIA~\cite{li2024aria} & DP & Causal & arXiv 2024 & \href{https://github.com/rhymes-ai/Aria}{\checkmark} \\

Qwen2-VL~\cite{wang2024qwen2vl} & DP & Causal & arXiv 2024 & \href{https://github.com/QwenLM/Qwen2-VL}{\checkmark} \\

Loong~\cite{wang2024loong} & VQ & Causal & arXiv 2024 & \ding{55} \\

Emu3~\cite{wang2024emu3} & VQ & Causal & arXiv 2024 & \href{https://github.com/baaivision/Emu3}{\checkmark} \\

MIO~\cite{wang2024mio} & LQ & Causal & arXiv 2024 & \ding{55} \\

G3PT~\cite{zhang2024g3pt} & VQ & Causal & arXiv 2024 & \ding{55} \\

Open-MAGVIT2~\cite{luo2024openmagvit2} & VQ & Causal & arXiv 2024 & \href{https://github.com/TencentARC/Open-MAGVIT2}{\checkmark} \\

VILA-U~\cite{wu2024vilau} & VQ & Causal & arXiv 2024 & \ding{55} \\

xGen-MM~\cite{xue2024xgenmm} & DP & Causal & arXiv 2024 & \href{https://github.com/salesforce/LAVIS/tree/xgen-mm}{\checkmark} \\

Lumina-mGPT~\cite{liu2024luminamgpt} & VQ & Causal & arXiv 2024 & \href{https://github.com/Alpha-VLLM/Lumina-mGPT}{\checkmark} \\

MARs~\cite{he2024mars} & VQ & Causal & arXiv 2024 & \href{https://github.com/fusiming3/MARS}{\checkmark} \\

OmniTokenizer~\cite{wang2024omnitokenizer} & VQ+LQ & Causal & arXiv 2024 & \href{https://github.com/FoundationVision/OmniTokenizer}{\checkmark} \\

GenLLaVA~\cite{hernandez2024genllava} & LQ & Causal & arXiv 2024 & \href{https://github.com/jeffhernandez1995/GenLlaVA}{\checkmark} \\

EVE~\cite{diao2024eve} & DP & Causal & arXiv 2024 & \href{https://github.com/baaivision/EVE}{\checkmark} \\

iVideoGPT~\cite{wu2024ivideogpt} & VQ & Causal & arXiv 2024 & \href{https://thuml.github.io/iVideoGPT/}{\checkmark} \\

InternVL2~\cite{chen2024internvl} & LQ & Causal & CVPR 2024 & \href{https://internvl.github.io/blog/2024-07-02-InternVL-2.0/}{\checkmark} \\

VisionLLM v2~\cite{wu2024visionllmv2} & LQ & Causal & NeurIPS 2024 & \href{https://github.com/OpenGVLab/VisionLLM/tree/main/VisionLLMv2}{\checkmark} \\

MAR~\cite{li2024mar} & LQ & Bidirectional & arXiv 2024 & \href{https://github.com/LTH14/mar}{\checkmark} \\

VAR~\cite{tian2024var} & VQ & Causal & NeurIPS 2024 & \href{https://github.com/FoundationVision/VAR}{\checkmark} \\

LlamaGen~\cite{sun2024llamagen} & VQ & Causal & arXiv 2024 & \href{https://github.com/FoundationVision/LlamaGen}{\checkmark} \\

Anole~\cite{chern2024anole} & VQ & Causal & arXiv 2024 & \href{https://github.com/GAIR-NLP/anole}{\checkmark} \\

Chameleon~\cite{team2024chameleon} & VQ & Causal & arXiv 2024 & \href{https://github.com/facebookresearch/chameleon}{\checkmark} \\

Libra~\cite{xulibra} & VQ & Causal & ICML 2024 & \href{https://github.com/YifanXu74/Libra}{\checkmark} \\

X-VILA~\cite{ye2024xvila} & LQ & Causal & arXiv 2024 & \ding{55} \\

WorldGPT~\cite{ge2024worldgpt} & DP & Causal & arXiv 2024 & \href{https://github.com/DCDmllm/WorldGPT}{\checkmark} \\

SEED-X~\cite{ge2024seedx} & LQ & Causal & arXiv 2024 & \href{https://github.com/AILab-CVC/SEED-X}{\checkmark} \\

GiT~\cite{wang2024git} & VQ & Causal & ECCV 2024 & \href{https://github.com/Haiyang-W/GiT}{\checkmark} \\

AnyGPT~\cite{zhan2024anygpt} & LQ & Causal & arXiv 2024 & \href{https://github.com/OpenMOSS/AnyGPT}{\checkmark} \\

DeLVM~\cite{hao2024delvm} & VQ & Causal & ICML 2024 & \href{https://github.com/ggjy/DeLVM}{\checkmark} \\

VideoLaVIT~\cite{jin2024videolavit} & VQ & Causal & ICML 2024 & \href{https://github.com/jy0205/LaVIT}{\checkmark} \\

AIM~\cite{el-nouby2024aim} & DP & Prefix & ICML 2024 & \href{https://github.com/apple/ml-aim}{\checkmark} \\

Unified-IO 2~\cite{lu2024unifiedio2} & VQ & Causal & CVPR 2024 & \href{https://github.com/allenai/unified-io-2}{\checkmark} \\

Emu2~\cite{sun2024emu2} & LQ & Causal & CVPR 2024 & \href{https://github.com/baaivision/Emu/tree/main/Emu2}{\checkmark} \\

VL-GPT~\cite{zhu2023vlgpt} & LQ & Causal & arXiv 2023 & \href{https://github.com/AILab-CVC/VL-GPT}{\checkmark} \\

LVM~\cite{bai2024lvm} & VQ & Causal & CVPR 2024 & \href{https://github.com/ytongbai/LVM}{\checkmark} \\

CoDi-2~\cite{tang2024codi2} & LQ & Causal & CVPR 2024 & \href{https://github.com/microsoft/i-Code/tree/main/CoDi-2}{\checkmark} \\

Mirasol3B~\cite{piergiovanni2024mirasol3b} & DP & Causal & CVPR 2024 & \href{https://github.com/kyegomez/Mirasol}{\checkmark} \\

TEAL~\cite{yang2023teal} & VQ & Causal & arXiv 2023 & \ding{55} \\

SEED-LLaMA~\cite{ge2024seedllama} & LQ & Causal & ICLR 2024 & \href{https://github.com/AILab-CVC/SEED}{\checkmark} \\

InstructSeq~\cite{fang2023instructseq} & VQ & Causal & arXiv 2023 & \href{https://github.com/rongyaofang/InstructSeq}{\checkmark} \\

LaViT~\cite{jin2024lavit} & VQ & Causal & ICLR 2024 & \href{https://github.com/jy0205/LaVIT}{\checkmark} \\

DreamLLM~\cite{dong2024dreamllm} & LQ & Causal & ICLR 2024 & \href{https://github.com/RunpeiDong/DreamLLM}{\checkmark} \\

Emu~\cite{sun2023emu} & LQ & Causal & ICLR 2024 & \href{https://github.com/baaivision/Emu}{\checkmark} \\

CM3Leon~\cite{yu2023cm3leon} & VQ & Causal & NeurIPS 2023 & \href{https://github.com/kyegomez/CM3Leon}{\checkmark} \\

GILL~\cite{koh2024gill} & LQ & Causal & NeurIPS 2023 & \href{https://github.com/kohjingyu/gill}{\checkmark} \\

Parti~\cite{yuparti} & VQ & Causal & TMLR 2022 & \href{https://github.com/google-research/parti}{\checkmark} \\

DALL-E~\cite{ramesh2021dalle} & VQ & Causal & ICML 2021 & \href{https://github.com/openai/DALL-E}{\checkmark} \\
\bottomrule
\end{tabular}

\caption{The overview of autoregressive vision foundation models. In the category of vision tokenizer, \textbf{VQ} denotes vector quantization, \textbf{DP} denotes direct projection, \textbf{LQ} denotes learnable query. In the category of autoregression backbone, \textbf{Causal} denotes causal Transformer ,\textbf{Bidirectional} denotes bidirectional Transformer, \textbf{Prefix} denotes prefix Transformer. \textbf{+} denotes both approaches are related.}
\label{tab:arvfm}
\end{table*}

\section{Methodology}\label{sec: 3_method}
As mentioned in Sec.~\ref{sec: 1_introduction}, the two key components of existing autoregressive vision foundation models are the vision tokenizer and autoregression backbone. The former provides a unified representation for inputs and outputs from the \textit{Space-Time-Modality} space, while the latter enables scalable processing of the data manifold. Based on the drawing style~\cite{wu2024cluster, li2024mar}, we present Fig.~\ref{fig:fig1} and Fig.~\ref{fig:fig2} according to our classification approach to illustrate the principles of different types of vision tokenizers and autoregression backbones, respectively. We also summarize autoregressive vision foundation models and label the types of their vision tokenizers and autoregression backbones in Tab.~\ref{tab:arvfm}, respectively.

\subsection{Vision Tokenizer}
The vision tokenizer aims to partition vision signals from the \textit{Space-Time-Modality} space, into a series of vision tokens, and then map them into vision embeddings that can be understood by the model. Based on the range of embedding values, vision tokenizers can be categorized into \emph{discrete-valued} and \emph{continuous-valued} approaches.

\subsubsection{Discrete-Valued Tokenizer}
\paragraph{Vector Quantization.} As shown in Fig.~\ref{fig:fig1} (a), Following NLP, a finite vocabulary or embedding space, \ie, codebook, is used to map words to fixed-dimensional and learnable vectors, \ie, codes, vision tokenizers can also employ codebooks to convert vision tokens into their corresponding codes, whether by mapping vision objects to concepts described by language words~\cite{peng2024visualtoken, wang2024git} or by directly using vision features to construct the codebook~\cite{van2017vqvae, esser2021vqgan}. Formally, given a codebook $B = \{b_1, b_2, \dots, b_K\} $, where $b_k$ represents the code, the mapping process is performed by finding the closest entry in the codebook for the token $x_i$:
\begin{equation}
    z_q(x_i) = b_{l_i}, \quad \text{where } l_i = \arg \min_{k} \| z_e(x_i) - b_k \|^2.
\end{equation}
Here, $z_e(x_i)$ is the embedding output by the vision encoder. $l_i$ is the index of the corresponding code of $z_e(x_i)$ in the codebook. $z_q(x_i)$ and $b_{l_i}$ are the quantized embedding, \ie, the selected code, indexed by $l_i$ in the codebook, which will later be fed into the vision decoder. Notably, $z_q(x_i)$ and $z_e(x_i)$ are continuous embeddings, while $l_i$ forms one of the elements of the discrete embedding $L = (l_1, l_2, \dots, l_K)$ for the image $x$. Van \etal~\cite{van2017vqvae} suggest that this approach can balance image generation quality with facilitating model learning. During inference, this discrete distribution can be sampled using a trained model, \eg, PixelCNN~\cite{van2016pixelcnn} and Transformer.

\paragraph{Highlight.} TEAL~\cite{yang2023teal}, OmniTokenizer~\cite{wang2024omnitokenizer} and InsturctSeq~\cite{fang2023instructseq} utilize VQ-VAE~\cite{van2017vqvae}, the first discrete vision tokenizer to introduce the codebook mechanism. DALL-E~\cite{ramesh2021dalle} employs dVAE, combining diverse codes through Gumbel-Softmax function rather than deterministically mapping to a single code. VILA-U~\cite{wu2024vilau} introduces RQ-VAE, which encodes the token by recursively stacking codes to increase representation capacity, thereby avoiding the computational overhead caused by enlarging the codebook. VQ-GAN~\cite{esser2021vqgan}, which improves reconstruction quality by introducing adversarial loss and perceptual loss, has been widely adopted by autoregressive vision foundation models~\cite{yuparti, wu2024janus, wang2024emu3, liu2024luminamgpt, he2024mars, sun2024llamagen, wang2024git, hao2024delvm, lu2024unifiedio2, bai2024lvm, wang2024loong}. Chameleon~\cite{team2024chameleon}, Anole~\cite{chern2024anole}, Libra~\cite{xulibra} and CM3Leon~\cite{yu2023cm3leon} deploy Make-A-Scene, which builds upon VQ-GAN by further adding scene layout control and perceptual loss for specific regions, \eg, salient objects. To reduce the significant computational overhead caused by traditional code lookup and comparison, the Lookup-Free Quantizer is recently introduced to compress the codebook into a set of integers and decompose the latent representation of each code into the cartesian product of single-dimensional variables~\cite{luo2024openmagvit2, zhang2024g3pt, wang2024loong}. Then, the input token is directly quantized using a sign function. LaViT~\cite{jin2024lavit} supports dynamically seleceting and merging tokens to reduce redundant information. VAR~\cite{tian2024var} turns the next token prediction to next scale prediction, which followed by HART~\cite{tang2024hart, liu2024sar}

\subsubsection{Continuous-Valued Tokenizer}
\begin{figure*}[!t]
  \includegraphics[width=1\textwidth]{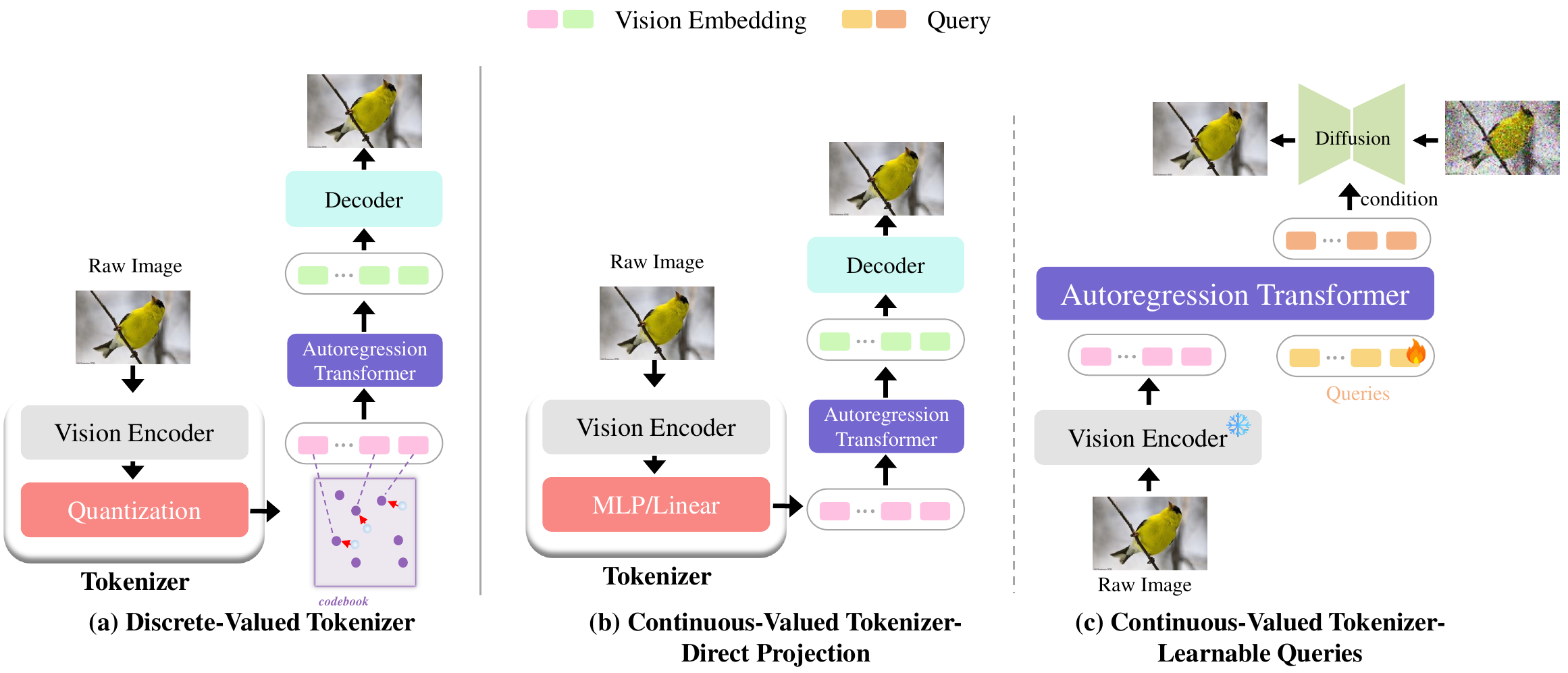}
  \caption{The overview of three types of vision tokenizers. (a) Discrete-valued tokenizer through vector quantization. (b) Continuous-valued tokenizer through direct projection. (c) Continuous-valued tokenizer through learnable queries.}
  \label{fig:fig1}
\end{figure*}

\paragraph{Direct Projection.} As shown in Fig.~\ref{fig:fig1} (b), some autoregressive vision foundation models~\cite{wu2024janus, li2024baichuanomni, li2024aria, wang2024qwen2vl, xue2024xgenmm, diao2024eve, ge2024worldgpt, el-nouby2024aim, piergiovanni2024mirasol3b} directly apply a projection module, \eg, MLP and Linear, or a pretrained vision encoder, \eg, EVA-CLIP~\cite{sun2023evaclip}, SigLIP~\cite{zhai2023siglip}, and CLIP, to map tokens to the continuous embeddings, which are subsequently mapped back to the image through a similar module. This process can be formulated as:
\begin{equation}
\begin{aligned}
    z_i &= \mathcal{E}_\theta(x_i), \\
    \hat{x}_i &= \mathcal{D}_\theta(\mathcal{F}_\theta(z_i)),
\end{aligned}
\end{equation}
where $\mathcal{E}_\theta(\cdot)$, $\mathcal{D}_\theta(\cdot)$ denotes the encoder and decoder, respectively. $\mathcal{F}_\theta$ denotes the backbone.
$z_i$ denotes the embedding projected by $\mathcal{E}_\theta(\cdot)$, and  $\hat{x}_i$ denotes the result reconstructed by the decoder.

\paragraph{Highlight.} Since the semantic granularity required for vision generation and understanding is inconsistent, Janus~\cite{wu2024janus} suggests that the vision tokenization should be decoupled into two ways: \textit{vector quantization}  for generation and \textit{direct projection} for understanding. Qwen2-VL~\cite{wang2024qwen2vl} integrates M-RoPE to address the issue of unified positional encoding across the \textit{Space-Time-Modality} space. EVE~\cite{diao2024eve} observes that pretrained vision encoder introduces inductive biases, \eg, inconsistent resolution and aspect ratio. The Patch Aligning Layer (PAL) is applied to align vision features generated by the model with pretrained vision encoder implicitly.

\paragraph{Learnable Query.} 
As shown in Fig.~\ref{fig:fig1} (c), in order to capture the potential cross-modal dependencies between the vision modality and other modalities, \eg, text, and adaptively extract the most relevant vision semantics, learnable queries are fed into the autoregressive vision foundation models. Given a set of learnable queries $Q = (q_1, q_2, \dots, q_L)$, where $q_i \in \mathbb{R}^d$ and $L$ denotes the length, assuming the position of special token used to indicate the start of the query is $j$, then the conditional embedding $C = (c_1, c_2, \dots, c_L)$ can be computed by:
\begin{equation}
    C = \mathcal{F}_\theta \left( Q, x_{<j+1} \right).
\end{equation}
$C$ is then used as the condition to the diffusion model for controllable decoding of the raw image. The entire process is categorized into two cases: the two-stage~\cite{koh2024gill, sun2023emu, ge2024seedllama, sun2024emu2, zhan2024anygpt, ge2024seedx, fang2024puma} and end-to-end~\cite{tang2024codi2, ye2024xvila, li2024mar, wu2024visionllmv2, chen2024internvl, wang2024omnitokenizer, wang2024mio, yang2024mmar, tang2024hart, fan2024fluid} training approaches. In the case of the two-stage approach, a vision encoder, \eg, EVA-CLIP, is first trained with the diffusion model to reconstruct the raw image. Then the pretrained vision encoder $\mathcal{E}_\phi$, is applied to generate the ground truth embedding $G = (g_1, g_2, \dots, g_L)$ of $C$ with the input of raw image $x$. This can be formalized as:
\begin{equation}
    G = \mathcal{E}_\phi(x).
\end{equation}
The embedding obtained from the autoregression backbone, \ie, conditional embedding $C$, is then supervised by $G$. $C$ is typically served as a condition for the diffusion model during inference. Instead of aligning $C$ and $G$ in the intermediate hidden states, the end-to-end approach directly applies generation loss of the diffusion model, \eg, score matching~\cite{songscore}, from the raw image space to optimize learnable queries, which are simultaneously incorporated into the diffusion model during training.

\paragraph{Highlight.} Fluid~\cite{fan2024fluid} has progressively demonstrated that the vision quality of images generated with random order and continuous-valued tokens is better compared to those with raster order and discrete-valued tokens. PUMA~\cite{fang2024puma} proposes a multi-granular vision feature decoding framework to better adapt the model to vision tasks at different levels. VisionLLM v2 introduces \textit{super link}, connecting adaptively the autoregression backbone with decoders for hundreds of vision tasks, enabling flexible routing and gradient transmit. X-VILA~\cite{ye2024xvila} introduces visual embedding highway, which helps preserve vision information during cross-modal alignment.

\subsection{Autoregression Backbone}
The autoregression backbone enables the autoregressive vision foundation model to process data in a streaming and scalable manner. Typically, backbones from LLMs are directly adopted, \eg, Llama~\cite{touvron2023llama}. The main differences lie in how they design Transformer architectures, and how vision tokens are fed into the model in a reasonable sequential pattern.

\subsubsection{Autoregression Transformer}
\begin{figure*}[!t]
\centering
\includegraphics[width=0.95\textwidth]{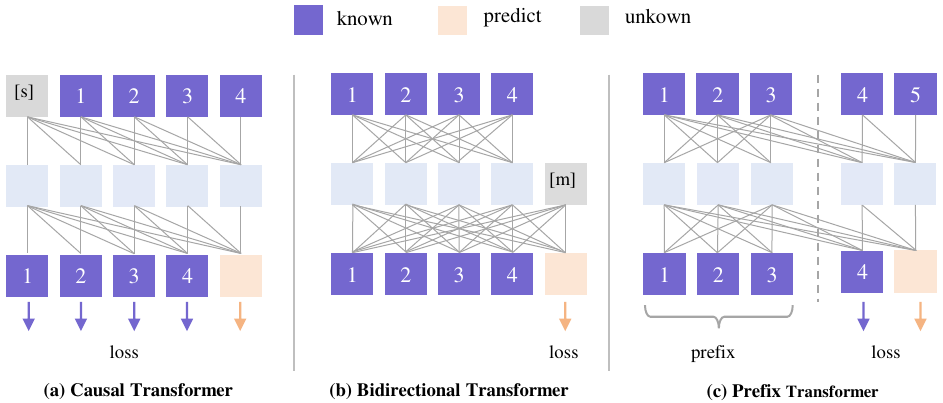}
\caption{The overview of Autoregression Architecture. (a) Causal attention in causal Transformer. (b) Bidirectional attention in bidirectional Transformer. (c) Prefix attention in prefix Transformer. The token with $\downarrow$ is included in the loss calculation.}
\label{fig:fig2}
\end{figure*}

\paragraph{Causal Transformer.} As shown in Fig.~\ref{fig:fig2} (a), the causal attention utilizes masks to ensure that each token can only attend to itself and the previous tokens in the sequence~\cite{fang2024puma, wu2024janus, tang2024hart, li2024baichuanomni, li2024aria, wang2024qwen2vl, wang2024loong, wang2024emu3, wang2024mio, zhang2024g3pt, luo2024openmagvit2, wu2024vilau, xue2024xgenmm, liu2024luminamgpt, he2024mars, wang2024omnitokenizer, hernandez2024genllava, diao2024eve, wu2024ivideogpt, chen2024internvl, wu2024visionllmv2, tian2024var, sun2024llamagen, chern2024anole, team2024chameleon, xulibra, ye2024xvila, ge2024worldgpt, ge2024seedx, wang2024git, zhan2024anygpt, hao2024delvm, jin2024videolavit, lu2024unifiedio2, sun2024emu2, zhu2023vlgpt, bai2024lvm, tang2024codi2, piergiovanni2024mirasol3b, yang2023teal, ge2024seedllama, fang2023instructseq, jin2024lavit, dong2024dreamllm, sun2023emu, yu2023cm3leon, koh2024gill, yuparti, ramesh2021dalle}. This restriction aligns intuitively with autoregression. Formally, the representation of the $i$-th token $y_i$ in a self-attention layer is computed as:
\begin{equation}
    y_i = \sum_{k=1}^{K} a_{ik} v_k, \label{selfattn}
\end{equation}
where $a_{ik}$ is the attention weight, indicating how much attention the $i$-th token gives to the $k$-th token. $v_k$ is the value embedding for the $k$-th token. In causal Transformer, masks are applied to attention weights, where $a_{ik} = 0 $ for $k > i$ and $\sum_{k=1}^{K} a_{ik} = 1$. During training, the input sequence is prepended with a start token $[s]$, and the model attempts to predict the corresponding sequence shifted one token to the left. The loss is computed for all tokens in the sequence:
\begin{equation}
    \min_{\theta} \frac{1}{K} \sum_{k=1}^{K} - \log P(x_k | x_{<k}, \theta).
\end{equation}
The next scale prediction proposed by VAR~\cite{tian2024var}, extending the causal mask from tokens to scales which can be viewed as images with different resolutions, \ie the scale to be predicted can only access the previously generated scales.

\paragraph{Bidirectional Transformer.} As shown in Fig.~\ref{fig:fig2} (b), the MAE-style autoregressive variant, named as MAR~\cite{li2024mar}, is proved that bidirectional attention can also perform autoregression~\cite{fan2024fluid, yang2024mmar}. Specifically, MAR reorders all tokens, and places unmasked tokens before the masked tokens. In this way, the masked tokens, \ie, the tokens to be predicted, are incidentally constrained to only access the previous tokens, as they are positioned at the end. Therefore, bidirectional attention allows each token to attend to all other tokens, and there is no need for causal masks to block attention weights. $a_{ik}$ in Eq. (\ref{selfattn}) satisfies $\sum_{k=1}^{K} a_{ik} = 1$. During training, masked tokens $[m]$ are added at the end in the middle layer, and are the only ones in the sequence for which the loss needs to be computed as:
\begin{equation}
    \min_{\theta} \frac{1}{|M|} \sum_{k \in M} - \log P(x_k | x_{<k}, \theta),
\end{equation}
where $M$ is the index set of masked tokens. Note that our analysis is based on the next token prediction. Models like MAR \etc can predict multiple masked tokens in parallel, but masked tokens are not visible to each other, regardless of whether they have already been predicted. They can only attend to previous unmasked tokens. SAR~\cite{liu2024sar} further removes this limitation by allowing unknown tokens, \ie, tokens waiting for prediction, to be dynamically divided into multiple sets and performing next set prediction, enabling unpredicted sets to observe predicted sets and the initially known tokens. They achieve this through a \textit{Fully Masked Transformer}, which can be viewed as a hybrid approach combining causal and bidirectional Transformer. 

\paragraph{Prefix Transformer.} As shown in Fig.~\ref{fig:fig2} (c), some models adopt prefix attention~\cite{el-nouby2024aim}, which also combines causal attention with bidirectional attention. In prefix Transformer, a prefix with the length of $S$ is treated with bidirectional attention, while the remainder of the sequence is processed with causal attention. For tokens in the prefix, \ie, $i \leq S$, $a_{ik}$ in Eq. (\ref{selfattn}) satisfies $a_{ik} = 0$ for $k > S$ and $\sum_{k=1}^{S} a_{ik} = 1$. For tokens outside the prefix, \ie, $i > S$, $a_{ik}$ in Eq. (\ref{selfattn}) satisfies $a_{ik} = 0$ for $k > i$ and $\sum_{k=1}^{K} a_{ik} = 1$. During training, only tokens beyond the prefix are computed for loss:
\begin{equation}
    \min_{\theta} \frac{1}{K-S} \sum_{k=S+1}^{K} - \log P(x_k | x_{<k}, \theta).
\end{equation}

\subsubsection{Autoregression Pattern}
By manipulating autoregression patterns, the model can access diverse manifolds of the input data, leading to more robust representations and improved learning dynamics. \textit{raster order}~\cite{el-nouby2024aim} means tokens are processed in the left-to-right or top-to-bottom order, which may be beneficial for structured inputs with a clear spatial and temporal orientation. \textit{spiral order}~\cite{el-nouby2024aim, liu2024sar} means tokens are processed in an outward spiral. This enables the model to focus on both local and global features, enhancing better context understanding. \textit{checkerboard order}~\cite{el-nouby2024aim} means tokens are processed in a staggered manner, alternating between tokens that are spatially distant. This allows the model to capture long-range dependencies earlier in the sequence. \textit{random order}~\cite{fan2024fluid} means the model is allowed to learn from unpredictable and varied token sequences. This approach can lead to better generalization and prevent overfitting to fixed token orders by introducing variability during training. \textit{clustered order} means tokens representing the same vision semantics are clustered together, and allowing for dynamically changing sequence lengths based on image complexity~\cite{wu2024cluster}. This helps preserve the integrity of vision semantic units, thereby preventing significant loss of high-frequency vision features. VAR~\cite{tian2024var} introduces \textit{next scale prediction}, \ie, \textit{scale order}, which predicts feature maps from coarse to fine granularity, and from low to high resolution, effectively preserving spatial relationships in images. SAR~\cite{liu2024sar} allows tokens to be divided into multiple sets, each with an arbitrary length, and predicts the next set step by step. Open-MAGVIT2~\cite{luo2024openmagvit2} divides the vocabulary into multiple sub-vocabularies. In this way, the latent representation of each token is spilled into several subspaces, \ie, each token consists of multiple subtokens. The model will then perform next subtoken prediction to avoid the high cost of updating the entire vocabulary at once. DnD-Transformer~\cite{chen2024dndtransformer} additionally extends autoregression along the model depth direction based on the 1D sequence. Matter \etal~\cite{mattar2024wavelets} decompose image details from coarse-grained to fine-grained, into wavelet coefficients.  

\section{Discussion}\label{sec: 4_discussion}
\subsection{Quality}
\paragraph{Discrete \textit{v.s.} Continuous.} Generally, discrete tokenization allows tokens to be associated with specific concepts, while continuous tokenization preserves more details. Recent studies~\cite{li2024mar, fan2024fluid, yang2024mmar} find that although both continuous and discrete tokenization satisfy the scaling law in terms of validation loss, continuous tokenization performs better in terms of generation quality, \eg, GenEval metric. However, recent discrete tokenization methods~\cite{yu2024magvit, luo2024openmagvit2, liu2024elm, hao2024bigr, weber2024maskbit} note that expanding the vocabulary can enhance reconstruction capability but may impair generation capability when achieving a certain scale, since the difficulty of token prediction increases accordingly. Therefore, they turn to look-up free techniques, which reduce the embedding dimension of the codebook and perform binary quantization on tokens.
Additionally, besides the stop gradient operation~\cite{van2017vqvae}, recent works~\cite{fifty2024rvae} further mitigate the information loss caused by the non-differentiability of discrete tokenization. In the future, discrete and continuous tokenization will remain in competition, while also presenting complementary opportunities, \eg, HART~\cite{tang2024hart} describes the overall image structure through discrete tokenization, while continuous tokenization is used to capture image details.  

\paragraph{AR \textit{v.s.} Diffusion.} Although autoregression and diffusion models are competing against each other intensely in the vision domain~\cite{tian2024var, sun2024llamagen}, we observe a trend towards their deep integration. In our survey, we have summarized that in continuous tokenization, the diffusion model receives learnable queries that capture historical semantics from the autoregression backbone to perform image decoding. Recently, Show-o~\cite{xie2024showo}, TransFusion~\cite{zhou2024transfusion} and MonoFormer~\cite{zhao2024monoformer} propose leveraging the advantages of autoregression for text and diffusion for images respectively. Furthermore, ACDC~\cite{chung2024acdc} points out that autoregression models are suitable for modeling long sequences but tend to accumulate exponential errors. In contrast, diffusion models are suitable for generating high-quality local contexts but are limited in generating long sequences. Therefore, they use a memory-conditioned diffusion model to refine the content generated by autoregression models locally. MovieDreamer~\cite{zhao2024moviedreamer} applies similar approaches to enhance the quality of video generation. There are also other cases of integration, \eg, generation diversity guidance~\cite{gu2024kaleidodiffusion} and various generation tasks~\cite{gu2024dart, xie2024progressive}. 

\paragraph{Unified and Controllable Modeling.} Existing autoregressive vision foundation models effectively unified modeling 2D images, then a significant trend is to extend them to 3D images~\cite{zhang2024g3pt, tang2024edgerunner, chen2024meshanythingv2, sun2024forest2seq} and videos~\cite{wang2024omnitokenizer, wu2024ivideogpt, jin2024videolavit, jin2024pyramidal}. Ultimately, an autoregressive vision foundation model unifying all understanding and generation tasks has the potential to serve as a powerful foundation to interact with the real world~\cite{zhang2024robot}. Although we can now unify all vision tasks through the visual sentence~\cite{bai2024lvm} to build a generalist vision foundation model, performance on most tasks still needs to be improved compared to specialized foundation models, even small models~\cite{ren2024mvg}. Therefore, in addition to advancing the scaling law, injecting conditional control into the deeper training process~\cite{yao2024car, li2024controlvar, li2024controlar} may also be helpful. Furthermore, we observe that the autoregression paradigm has been continuously optimized, \eg, from VQ-VAE~\cite{van2017vqvae} to VAR~\cite{tian2024var}. Exploring autoregressive patterns by leveraging intrinsic characteristics of images, \eg, medical images, 3D images, and videos, is worthwhile. In the future, autoregressive vision foundation models that completely cover the \textit{Space-Time-Modality} space, more optimized autoregression pattern and more flexible conditional control methods~\cite{zhang2024varclip, ma2024star} are also promising research directions.

\subsection{Efficiency}
A limitation of autoregression models is their substantial time consumption during inference~\cite{chen2024understanding,xia2024unlocking}. This issue arises from their sequential nature, which necessitates hundreds or even thousands of steps to predict one token at a time. Despite the commonly used training/inference acceleration techniques for general foundation models~\cite{hooper2024kvquant, dao2022flashattention, daoflashattention2}, a intuitive strategy to overcome this limitation is to enable autoregression models to decode multiple tokens simultaneously within a single forward pass~\cite{gloeckle2024better,teng2024accelerating, li2024mar}. For example, Teng \etal~\cite{teng2024accelerating} propose a training-free probabilistic parallel decoding algorithm called \emph{Speculative Jacobi Decoding} (SJD). This method allows the model to predict multiple tokens at each step and accepts tokens based on a probabilistic criterion. This approach contrasts with the original Jacobi decoding, which uses a slower deterministic criterion to determine convergence, thereby accelerating the model by reducing the number of steps needed compared to the conventional next-token-prediction paradigm. More recently, Jang \etal~\cite{jang2024lantern} examine the token selection ambiguity, \ie, autoregression vision models often assign uniformly low probabilities to tokens, thereby hampering the performance of speculative decoding. They propose a relaxed acceptance condition that leverages the interchangeability of tokens in the latent space to make a more flexible use of candidate tokens. Yu \etal~\cite{yu2024titok} propose TiTok, which suggests that directly downsampling 2D images to obtain latent representations cannot effectively model the relationships between adjacent regions of the image. Instead, they use an additional smaller set of latent codes to represent the image. In particular, when dealing with 3D images, videos, and high-resolution 2D images, \eg, pathological slices, the number of vision tokens will grow exponentially. At this time, leveraging vision priors to decouple tokens or utilize redundant tokens dynamically is also a promising direction~\cite{jin2024videolavit, jin2024lavit, chen2024imageplug, koner2024lookupvit, xu2024gigapath}.  

\subsection{Evaluation}
Fair and easy-to-use benchmarks ensure the consistency and fairness of evaluations, reducing time spent on reproducing baselines ~\cite{kang2023studiogan}. This not only signifies maturity but also greatly advances the progress, as has been validated by ample work ~\cite{yue2024mmmu,li2024seed,zhong2023agieval,wu2023q,bai2024benchmarking}. These successful benchmarks filter promising methods and possess common characteristics: continuously growing data scales~\cite{zhu2024genimage}, widely adopted evaluation metrics and tasks ~\cite{yu2023mm}, and a combination of both quantity and quality continuously updated SOTA methods ~\cite{sun2024journeydb}.

Benchmarks for image generation are particularly complex ~\cite{bitton2023breaking, zhu2024genimage}. Apart from the obvious time-consuming nature of image generation ~\cite{jang2024lantern}, generating more realistic images and improving the detection accuracy of generated images ~\cite{zhu2024genimage} add to the difficulty of image generation benchmarks. This challenge also applies to autoregressive vision foundation models. In recent years, significant methods have emerged ~\cite{tian2024visual,sun2024llamagen}, although they have often focused on specific tasks, such as evaluating the quality of generated objects using metrics like the Inception Score (IS) and Fréchet Inception Distance (FID) \etc~However, efforts are being made to achieve the aforementioned characteristics and to capture the multifaceted capabilities required by vision foundation models, which are expected to handle a variety of multimodal tasks, such as understanding and generation.

Therefore, it is imperative to develop fair and consistent benchmarks to address the challenges mentioned above in autoregressive image generation. On the other hand, for the integration of understanding and generation, we are witnessing the emergence of more targeted benchmarks ~\cite{zhou2024vega,yu2024mm}. The design of these interleaved datasets represents a promising direction. In the future, we believe that more comprehensive datasets and benchmarks will be constructed to cover the \textit{Space-Time-Modality} space. In addition to evaluating improvements in generation quality and efficiency, further assessments can focus on other challenging issues, \eg, hallucination, mathematical reasoning, and long-context learning.

\section{Conclusion}\label{sec: 5_conclusion}
In this paper, we conduct a comprehensive survey of autoregressive vision foundation models by offering an in-depth analysis of their vision tokenizers, autoregression backbones. We also present potential future directions for autoregressive vision foundation models. To the best of our knowledge, this is the first survey to comprehensively summarize autoregressive vision foundation models under the trend of unifying understanding and generation. We hope this survey will inspire researchers interested in autoregressive vision foundation models, and provide insights on how to better advance the unification of vision understanding and generation in the future.

\bibliography{ref}
\bibliographystyle{unsrt}

\clearpage
\appendix
\onecolumn

\end{document}